\documentclass{article}
\usepackage{PRIMEarxiv}

\usepackage[utf8]{inputenc} 
\usepackage[T1]{fontenc}    
\usepackage{hyperref}       
\usepackage{url}            
\usepackage{booktabs}       
\usepackage{amsfonts}       
\usepackage{nicefrac}       
\usepackage{microtype}      
\usepackage{lipsum}
\usepackage{fancyhdr}       
\usepackage{graphicx}       
\graphicspath{{media/}}     
\usepackage{natbib}
\usepackage{amsmath}
\usepackage{amssymb}
\usepackage{cleveref}
\usepackage[official]{eurosym}
\usepackage{algorithm}
\usepackage{algpseudocodex}
\usepackage{float}
\usepackage{array,tabularx}
\usepackage{booktabs} 
\usepackage{makeidx}
\usepackage{makecell}
\usepackage{xparse}
\usepackage{moredefs}
\usepackage{lips}
\usepackage{xcolor}

\title{A Memetic NSGA-III for Green Flexible Production with Real-Time Energy Costs \& Emissions
\thanks{\textit{\underline{Citation}}: 
\textbf{Burmeister, S.C., 2025. A Memetic NSGA-III for Green Flexible Production with Real-Time Energy Costs \& Emissions. Croatian Operational Research Review, 16(2). DOI:\href{https://doi.org/10.17535/crorr.2025.0009}{10.17535/crorr.2025.0009}.}} 
}

\author{
  Sascha C Burmeister \\
  Management Information Systems \\
  Paderborn University \\
  Paderborn, Germany\\
  sascha.burmeister@upb.de\\
}

\begin{document}
\maketitle

\begin{abstract}
The use of renewable energies strengthens decarbonization strategies. 
To integrate volatile renewable sources, energy systems require grid expansion, storage capabilities, or flexible consumption.
This study focuses on industries that adapt production to real-time energy markets, offering flexible consumption to the grid. 
Flexible production considers not only traditional goals like minimizing production time, but also minimizing energy costs and emissions, thereby enhancing the sustainability of businesses.
However, existing research focuses on single goals, neglects the combination of makespan, energy costs, and emissions, or assumes constant or periodic tariffs instead of a dynamic energy market.
We present a novel memetic NSGA-III to minimize makespan, energy cost, and emissions, integrating real energy market data, and allowing manufacturers to adapt energy consumption to current grid conditions.
Evaluating it with benchmark instances from literature and real energy market data, we explore the trade-offs between objectives, showcasing potential savings in energy costs and emissions on estimated Pareto fronts.
\end{abstract}


\section{Introduction}
Renewable energy is a key solution to the challenge of climate change.
However, traditional energy systems face high penetration of volatile renewable energy sources and face new challenges in terms of power quality, reliability, or power system reliability~\citep{basit2020limitations}.
To support the integration of renewable energy sources, energy systems need grid expansion, storage capabilities, or flexible loads.
In the context of flexible loads, consumers are encouraged to shift energy demand away from peak times to relieve grid load and toward periods of high renewable generation, promoting sustainable energy consumption.
For manufacturing industries as large energy consumers, flexible loads offer a high potential for cost savings~\citep{keller2017integration}.
Additionally, the sustainability of energy production can also be considered.
Instead of pricing emitted emissions through CO\textsubscript{2} taxes and thus implicitly incorporating them into energy costs, it can be beneficial to explicitly account for emissions in production planning.
This allows focused attention on a company's carbon footprint, which is advantageous for reasons such as anticipating or responding to regulatory pressures, attracting environmentally conscious customers, and maintaining legitimacy with external stakeholders~\citep{dahlmann2019managing}.
This work focuses on the challenge manufacturers face in aligning their production schedule with fluctuations in energy prices and emissions of the energy market.

In the energy market, there are three classic types of energy tariffs: (1) fixed energy tariffs with constant energy prices, (2) time-of-use (TOU) tariffs, where energy prices are divided into several time periods (e.g., on-peak and off-peak prices), and (3) real-time pricing (RTP) tariffs, where the cost of electricity consumption is based on the electricity exchange and changes at least every hour.
The latter allows manufacturers to align their production with the energy market, respond to price signals, adapt flexibly to the needs of the power system, and thereby reduce energy cost and emissions~\citep{finn2014demand}.

In the field of operations research, classical scheduling problems prioritize economic factors such as the schedule's makespan.
Green Job Shop Scheduling Problems are an extension of classical scheduling problems, incorporating resource and environmental aspects~\citep{li2022review}.
As we show in Section~\ref{sec:literature}, recent research places different focuses on economic goals such as the makespan or energy costs and ecological goals such as emissions.
To the best of our knowledge, no research has combined the minimization of makespan, energy cost, and emissions while considering RTP tariffs.
However, integrating these factors is essential for production planners to accurately calculate the potential reductions in energy costs and emissions achievable through improved production flexibility.

In this work, we focus on the effects of flexible production by exploring the research question:
How do scheduling decisions that prioritize one objective — makespan, energy cost, or emissions — affect the others?
Our contributions include (1) the formulation of a Flexible Job Shop Scheduling Problem (FJSP) with dynamic energy cost and emissions, (2) developing a memetic algorithm based on the Non-Dominated Sorting Algorithm III (NSGA-III), and (3) conducting computational experiments to evaluate the trade-offs among makespan, energy costs, and emissions.
In our study, we continue the work of~\cite{burmeister2023memetic}, who formulate a model for the bicriteria FJSP with respect to makespan and energy cost and present a memetic NSGA-II as a solution approach.
Our mathematical model extends this framework to include emissions as a third objective.
Additionally, our memetic NSGA-III enhances the existing memetic NSGA-II, enabling the calculation of multi-objective schedules aimed at minimizing makespan, energy cost, and emissions.
This extension allows us to evaluate trade-offs among all three objectives in our computational experiments and quantify potential savings for practitioners.

As a solution approach, we opt for memetic NSGA-III due to its ability to represent schedules in a three-dimensional Pareto front, allowing decision-makers to balance trade-offs and select solutions based on their preferences.
It also leads to favorable results in similar FJSP problems~\citep{sang2022intelligent,wu2021adaptive} and has already been used to optimize schedules with respect to energy-related goals~\citep{sun2021modified}.
A notable aspect of our study is the evaluation of the scheduling problem using real energy costs and emissions data from the German energy market.

The remainder of this paper is organized as follows.
Section~\ref{sec:literature} presents recent research in the area of green scheduling.
Section~\ref{sec:model} introduces the mathematical model, for which we present a memetic NSGA-III in Section~\ref{sec:algorithm}.
Section~\ref{sec:compex} shows our computational experiments and discusses our results.
Section~\ref{sec:conclusion} summarizes our results and recommends directions for future research. 

\section{Recent research} \label{sec:literature}
In this section, we present research related to energy cost- and emission-aware scheduling.
We discuss the respective objectives and solution approaches presented in the literature.

A schedule can be designed to minimize makespan, emissions, energy cost, or a combination of these.
In order to take ecological and economic goals into account, one approach is to minimize makespan while limiting energy consumption~\citep{carlucci2021job}.
Energy consumption can also be minimized as an objective, but is still independent of the underlying energy mix and its costs and emissions~\citep{lu2021knowledge,sun2021modified}.
\citet{wang2020multi} incorporate an economic view of energy consumption and minimize the makespan and energy costs.
They divide a day into several periods with different energy prices based on a TOU tariff.
A more detailed consideration of dynamic prices with RTP tariffs can be found in \citet{abikarram2019energy} and \citet{fazli2018energy}.
They consider hourly-changing prices but neglect a minimization of the emissions and the makespan of the production schedule.
\citet{burmeister2023memetic} present a model for bicriteria optimization of makespan and energy cost, considering an RTP tariff.

Due to the NP-hard complexity of Job Shop Scheduling Problems~\citep{garey1976complexity}, many studies favor metaheuristics over exact solution methods, allowing fast adaptation to fluctuating prices and emissions.
\citet{schulz2019multi} use a multiphase iterated local search algorithm to determine a Pareto front regarding makespan, total energy costs and peak load.
Most studies on energy-efficient scheduling reviewed in \citet{gao2020review} employ swarm intelligence and evolutionary algorithms.
\citet{dong2022green} design an improved hybrid salp swarm and NSGA-III algorithm to reduce carbon emissions and energy costs under the TOU tariff.
\citet{lu2021knowledge} and \citet{wang2020multi} apply a multi-objective Genetic Algorithm, while \citet{burmeister2023memetic} and \citet{sun2021modified} use multi-objective evolutionary metaheuristics for fast computation.

Recent research has focused on several areas of green scheduling.
However, the reviewed works that include energy costs and emissions neglect the real-time energy market, while studies related to real-time energy markets do not include emissions.
We aim to fill this research gap by combining economic and environmental perspectives, taking into account a dynamic energy market.

\section{Mathematical model}\label{sec:model}
In this section, we present the mathematical optimization model for the multi-objective FJSP with the objectives of minimizing makespan, energy cost, and emissions.
The model is based on the research conducted by~\cite{burmeister2023memetic}, which we extend to take into account emissions.
Table \ref{tab:notation} presents the notation for the mathematical model.
The set $J = \{1,...,\mu\}$ contains $\mu$ jobs to be processed, each of which is divided into $v_i$ operations $O_i=\{(i,1),...,(i,v_i)\}$. 
The operations of a job must be processed in sequence. 
The set $O=\bigcup_{i\in J} O_i$ contains all operations.
For each operation, $\tau_{ijk}$ specifies the duration with which an operation $(i,j)$ can be processed on machine $k\in M$. 
Supplementary, $\eta_{ijkt}$ and $\zeta_{ijkt}$ contain the energy cost and emissions for processing operation $(i,j)$ on machine $k\in M$ beginning at time $t\in T$.

\begin{table}[h]
	\centering
	\caption{Notation for the math. formulation}\label{tab:notation}
	\begin{tabular}{lp{4.6cm}|lp{4.6cm}}
		\toprule
        \bf Notation & \bf Description & \bf Notation & \bf Description\\ 
        \midrule
		\multicolumn{2}{l|}{\textit{Sets}} & \multicolumn{2}{l}{\textit{Variables}}\\
		$J$ & Jobs, $i \in J$ &
            $c^{max}$ & Maximum makespan\\
		$O$ & Operations, $O=\bigcup\limits_{i\in J} O_i$, &
            $p^{sum}$ & Sum of all energy cost\\
        & $O_i=\{(i,1),...,(i,\nu_i)\}$ &
            $e^{sum}$ & Sum of all emissions\\
		$M$ & Machines, $k\in M$&
            $s_{ijk}$ & Start time of operation $(i,j)$ \\
		$T$ & Time steps, $t\in T$&
            &on machine~$k$\\
        \multicolumn{2}{l|}{\textit{Parameters}} & 
            $c_{ijk}$ & End time of operation $(i,j)$\\
        $L$ & A large number &
            &on machine~$k$\\
		$\tau_{ijk}$ & Processing time of operation $(i,j)$ on machine $k$ &
            $x_{ijk}$ & Binary indicator, 1 iff operation $(i,j)$ is allocated on machine $k$\\
        $\eta_{ijkt}$ & Energy cost for processing operation $(i,j)$ on machine $k$ when starting at time~$t$ &
            $y_{iji'j'k}$ & Binary indicator, 1 iff operation $(i,j)$ is predecessor of operation $(i',j')$ on machine $k$\\
        $\zeta_{ijkt}$ & Energy emissions for processing operation $(i,j)$ on machine $k$ when starting at time $t$ &
            $p_{ijkt}$ & Binary indicator, 1 iff operation $(i,j)$ starts on machine $k$ at time~$t$\\
		\bottomrule
	\end{tabular}
\end{table}

Objective function \ref{eq:obj} minimizes the variables $c^{max}$, $p^{sum}$, and $e^{sum}$, which represent the maximum makespan, the sum of all energy cost, and the sum of all emissions, respectively.
Constraint \ref{eq:makespan} reflects the final completion time across all operations $(i,j)\in O$ and machines $k\in M$.
Constraints \ref{eq:cost} and \ref{eq:emissions} sum the energy cost $\eta_{ijkt}$ and emissions $\zeta_{ijkt}$, respectively, for all operations $(i,j)\in O$ on all machines $k\in M$ at all time steps $t\in T$.

Constraints \ref{eq:job-allocation} to \ref{eq:machine-predecessor-2} are based on the MILP formulation for the general FJSP of~\cite{ozguven2010mathematical}.
They ensure that each operation is assigned to exactly one machine (Constraint \ref{eq:job-allocation}), that operations of a job can only start if previously required operations have been completed (Constraints \ref{eq:relation-job-alloc-start-end} to \ref{eq:job-start}), and that operations of a machine cannot overlap (Constraints \ref{eq:machine-predecessor-1} and \ref{eq:machine-predecessor-2}).

To consider the energy consumption in the model, we add a link from the allocation of the operations to their individual energy consumption:
Constraint \ref{eq:link-x-and-p-1} forces the sum over binary indicators $p_{ijkt}$ over all time steps $t$ to be one, if operation $(i,j)$ is assigned to machine $k$.
Constraints \ref{eq:link-x-and-p-2} and \ref{eq:link-x-and-p-3} ensure that the binary indicator $p_{ijkt}$ is set to one for the time $t$ at which the operation $(i,j)$ starts.

The model is composed of binary and continuous variables. 
Together with convex and linear constraints, it is classified as an MILP.

\let\displaystyle\textstyle
\begin{align}
	\text{min } &(c^{max}, p^{sum}, e^{sum}) 	& \label{eq:obj} \\
	\text{s.t. }&c^{max} \geq c_{ijk} 			& \forall i,j,k\label{eq:makespan} \\
    &p^{sum} \geq \sum_{i,j,k,t} \eta_{ijkt} p_{ijkt} & \label{eq:cost}\\
	&e^{sum} \geq \sum_{i,j,k,t} \zeta_{ijkt} p_{ijkt} & \label{eq:emissions}\\
    &\sum_k x_{ijk} = 1 						& \forall i,j \label{eq:job-allocation} &\\
    &s_{ijk} + c_{ijk} \leq x_{ijk}L			& \forall i,j,k\label{eq:relation-job-alloc-start-end}\\
    &c_{ijk} \geq s_{ijk} + \tau_{ijk} - (1-x_{ijk})L & \forall i,j,k\label{eq:operation-process-time} \\
	&\sum_k s_{ijk}\geq\sum_k c_{i,j-1,k} 		& \forall i,j\label{eq:job-start} \\
    &s_{ijk} \geq c_{i'j'k} - y_{iji'j'k}L 		& \forall i,j,i',j',k\label{eq:machine-predecessor-1}\\
	&s_{i'j'k} \geq c_{ijk} - (1-y_{iji'j'k})L 	& \forall i,j,i',j',k\label{eq:machine-predecessor-2} \\
	&x_{ijk} = \sum_t p_{ijkt} 					& \forall i,j,k\label{eq:link-x-and-p-1} \\
	&s_{ijk} - t \geq -(1-p_{ijkt})L 			& \forall i,j,k,t\label{eq:link-x-and-p-2} \\
	&s_{ijk} - t \leq (1-p_{ijkt})L 			& \forall i,j,k,t\label{eq:link-x-and-p-3}\\
	&c^{max}, p^{sum}, e^{sum}, s_{ijk}, c_{ijk} \in \mathbb{R}_+ ,  \nonumber&\\
	&x_{ijk}, y_{iji'j'k}, p_{ijkt} \in \{0,1\} \hspace{1.6em} \forall i,j,i',j',k,t	&	
\end{align}

\section{The memetic NSGA-III}\label{sec:algorithm}
In this section, we present the memetic NSGA-III for solving the green multi-objective FJSP.
We divide the section into two parts, first introducing the representation of solutions in genotypes and phenotypes in Subsection~\ref{ssec:algorithm-representation}, before explaining the algorithm of the memetic NSGA-III in Subsection~\ref{ssec:algorithm-code}.

\subsection{Representation of solutions} \label{ssec:algorithm-representation}
The NSGA-III is an evolutionary algorithm and describes solutions as individuals of a population that evolve over multiple generations.
To represent solutions as individuals, we follow a decoder-based approach that encodes solutions as genotypes and uses phenotypes for decoding.
We base our genotype on the work of \cite{dai2019multi}, who choose a bipartite gene string for the sequence of operations and their machine assignment, and \cite{burmeister2023memetic}, who choose a tripartite gene string for the additional representation of the maximum allowable energy cost per operation.
To account for emissions, we extend latter approach and add a fourth gene string to this representation to indicate the maximum allowable emissions per operation.
Fig.~\ref{fig:genotype} shows an example genotype for three jobs to be assigned to two machines.

\begin{figure}[ht]
    \centering
	\includegraphics[width=.8\textwidth]{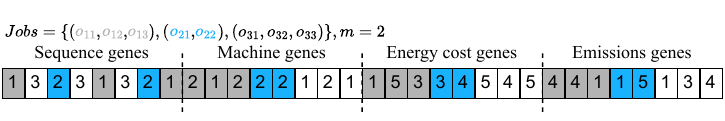}
	\caption{Example genotype for solution encoding}
	\label{fig:genotype}
\end{figure}

Fig.~\ref{fig:phenotype} illustrates the representation of the genotype on Fig.~\ref{fig:genotype}.
The allocation of operations to the machines is plotted as a Gantt chart, with time steps noted on the x-axis and machines on the left y-axis.
On the right y-axis are the energy cost values, shown as a dashed line, and the emissions values, shown as a dotted line.
Energy cost are given in \euro/MWh and emissions are given in grams of carbon dioxide equivalent (gCO\textsubscript{2}eq) per kWh.
The sequence gene string specifies the order in which the jobs are placed, while the machine gene string reflects the machine to be selected.
Thus, operation (1,1) is first assigned to machine 2.
The operation is scheduled at the first possible time that satisfies both the associated maximum allowable energy cost of the energy cost gene string and the emissions of the emissions gene string.
This means that operation (1,1) is scheduled at the first time that costs less than or equal to 1~\euro{} and less than or equal to 4~gCO$_2$eq, which is time step 4.

\begin{figure}[ht]
    \centering
	\includegraphics[width=.6\textwidth]{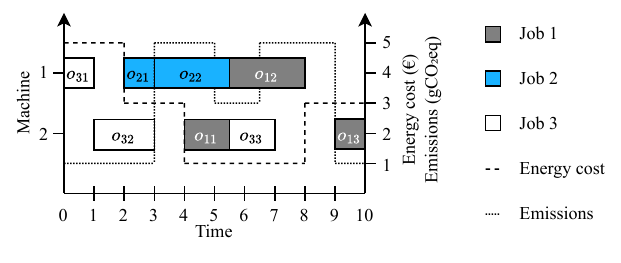}
	\caption{Representation of the example genotype as a phenotype}
	\label{fig:phenotype}
\end{figure}

All jobs are scheduled in the order specified by the sequence gene string.
If the time horizon considered is not sufficient to place an operation, e.g., because the allowed energy cost or emissions are too high, either the time horizon can be extended or the energy cost and emissions can be reduced.
While the former tends to produce long schedules with favorable energy cost and emissions, the latter tends to produce fast and expensive schedules.
To avoid bias and promote population diversification, we alternate the approach in each generation.
Overall, the genotype and phenotype can represent any sequence of operations on any combination of machines, energy cost, and emissions, resulting in a complete representation of the solution space.

\subsection{Algorithm} \label{ssec:algorithm-code}

In this section, we present the memetic NSGA-III for solving the green multi-objective FJSP.
We divide the section into two parts, first introducing the representation of solutions in genotypes and phenotypes in Section~\ref{ssec:algorithm-representation}, before explaining the algorithm of memetic NSGA-III in Section~\ref{ssec:algorithm-code}.

\subsection{Representation of solutions} \label{ssec:algorithm-representation}
The NSGA-III is an evolutionary algorithm and represents solutions as individuals of a population that evolve over multiple generations.
To represent solutions as individuals, we follow a decoder-based approach that encodes solutions as genotypes and uses phenotypes for decoding.
We base our genotype on the work of~\cite{dai2019multi}, who choose a bipartite gene string for the sequence of operations and their machine assignment, and~\cite{burmeister2023memetic}, who choose a tripartite gene string for the additional representation of the maximum allowable energy cost per operation.
To account for emissions, we extend the latter approach and add a fourth gene string to this representation to indicate the maximum allowable emissions per operation.

\begin{figure}[b]
    \centering
	\includegraphics[width=\textwidth]{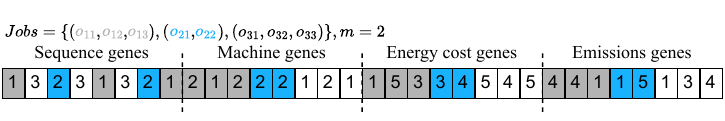}
	\caption{Example genotype for solution encoding based on~\cite{burmeister2023memetic}}
	\label{fig:genotype}
\end{figure}

Figure~\ref{fig:genotype} shows an example genotype for three jobs to be assigned to two machines.
The genotype is divided into four strings of genes, representing the sequence in which operations are allocated, the machine allocation, the tolerated energy costs, and the tolerated emissions.
Genes linked to jobs 1, 2, and 3 are highlighted in gray, blue, and white, respectively.
Each section has a length corresponding to the total number of operations.
For decoding the genotype in Figure~\ref{fig:genotype} as a schedule, Figure~\ref{fig:phenotype} illustrates the phenotype.
The allocation of operations to the machines is plotted as a Gantt chart, with time steps noted on the x-axis and machines on the left y-axis.
On the right y-axis are the energy cost values, shown as a dashed line, and the emissions values, shown as a dotted line.
Energy cost is given in \euro/MWh and emissions are given in grams of carbon dioxide equivalent (gCO\textsubscript{2}eq) per kWh.
The sequence gene string specifies the order in which the jobs are placed, while the machine gene string reflects the machine to be selected.
Thus, operation (1,1) is first assigned to machine 2.
The operation is scheduled at the first possible time that satisfies both the associated maximum allowable energy cost of the energy cost gene string and the emissions of the emissions gene string.
This means that operation (1,1) is scheduled at the first time that costs less than or equal to 1~\euro{} and less than or equal to 4~gCO$_2$eq, which is time step 4.

\begin{figure}[ht]
    \centering
	\includegraphics[width=.8\textwidth]{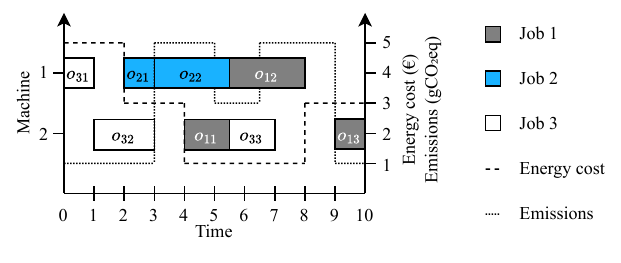}
	\caption{Representation of the example genotype as a phenotype based on~\cite{burmeister2023memetic}}
	\label{fig:phenotype}
\end{figure}

All jobs are scheduled in the order specified by the sequence gene string.
If the time horizon considered is not sufficient to place an operation, e.g., because the allowed energy cost or emissions are too high, either the time horizon can be extended or the energy cost and emissions can be reduced.
While the former tends to produce long schedules with favorable energy cost and emissions, the latter tends to produce fast and expensive schedules.
To avoid bias and promote population diversification, we alternate the approach in each generation.
Overall, the genotype and phenotype can represent any sequence of operations on any combination of machines, energy cost, and emissions, resulting in a complete representation of the solution space.

\subsection{Algorithm} \label{ssec:algorithm-code}

In this section, we explain our memetic NSGA-III.
We base the algorithm on the NSGA-III~\citep{deb2013evolutionary}, which uses a similar framework to its predecessor NSGA-II~\citep{deb2002fast} while providing better diversity.

The NSGA-III is outlined in \Cref{alg:nsga3}.
For a population $P_t$ of size $N$, NSGA-III generates an equally sized next generation $P_{t+1}$ using recombination and mutation operators:
For recombination, we use a two-point crossover to intensify the search in the solution space and to find better individuals based on existing ones.
From two parents, the genes of the gene strings are swapped between two different random positions. 
The same positions for swapping are chosen for all gene strings. 
By swapping, the gene strings remain feasible for machine assignment, energy cost, and emissions.
In the case of the sequence gene string, swapping may cause an infeasibility, e.g., if a job is no longer listed in the number of its operations. 
In this case, the defective children are repaired by replacing excess jobs with missing ones.
For mutation, the algorithm randomly modifies gene strings by performing one of the following changes: swapping two genes within the sequence gene string, reassigning a gene in the machine gene string, or assigning new values to either the energy cost gene string or the emission gene string.

\begin{algorithm}
	\caption{Pseudocode of the memetic NSGA-III based on~\cite{deb2013evolutionary}} \label{alg:nsga3}
	\begin{algorithmic}
		\Require Population $P_t$, Population size $N$
		\State $S_t = \emptyset, i=1$
		\State $R_t \gets P_t \cup \Call{Recombination+Mutation}{P_t}$
		\State $R_t \gets \Call{LocalRefinement}{R_t}$
		\State $(F_1,F_2,...) =  \Call{Non-dominated-sort}{R_t}$
		\Repeat
			\State $S_t = S_t \cup F_i$ and $i \gets i+1$
		\Until $\vert S_t\vert \geq N$
		\If{$\vert S_t\vert = N$}
			\State $P_{t+1} \gets S_t$
		\Else			
			\State $P_{t+1} \gets \bigcup^{i-1}_{j=1}F_j$
			\State $k \gets N-\vert P_{t+1}\vert$ 
			\Comment{Remaining space in $P_{t+1}$}
			\State $Z^r \gets \Call{Normalization}{S_t}$
			\State $\Call{Association}{Z^r,S_t}$
			\State $P_{t+1}\gets P_{t+1}\cup \Call{Niching}{F_i, k}$
		\EndIf
	\end{algorithmic}
\end{algorithm}

\begin{figure}[h]
    \centering
	\includegraphics[width=.8\linewidth]{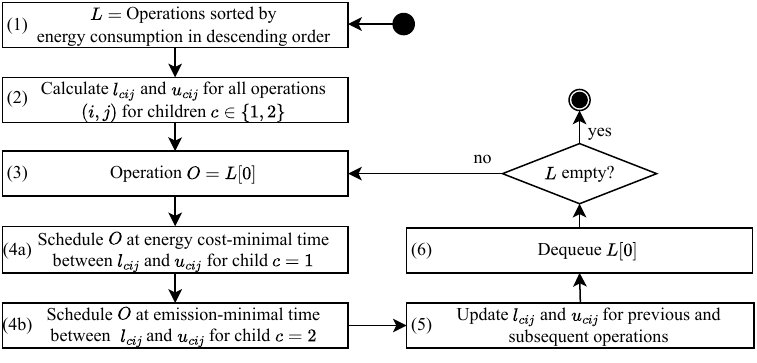}
	\caption{Greedy local refinement}
	\label{fig:greedy}
\end{figure}

We extend NSGA-III to a memetic algorithm and adopt local refinement from \citet{burmeister2023memetic}.
In their local refinement, a greedy approach improves solutions by adjusting the values of the gene strings for energy cost, ensuring that lower energy cost are achieved without increasing the makespan.
We extend this approach to include emissions as shown in Figure~\ref{fig:greedy}.
(1)~First, it sorts all operations of the parent, based on their energy consumption, in descending order into a queue $L$. 
(2)~Second, it calculates the lower and upper feasible start times $l_{cij}$ and $u_{cij}$ for each operation $(i,j)\in O$ based on the duration of previous and subsequent operations of the job for both children $c\in\{1,2\}$. 
(3)~Third, the greedy procedure selects the operation with the highest energy consumption, and (4)~fourth, schedules the selected operation at the time of the cheapest energy cost (4a) and emissions (4b), respectively.
(5)~Fifth, it adjusts the earliest possible start and end times of the operations that are still to be sorted, and (6) dequeues the current operation from $L$.
The procedure repeats steps (3) and (4) successively for the next operation until all operations in $L$ have been scheduled.

After recombination, mutation, and local refinement, the resulting set $R_t$ is sorted into several non-dominated fronts as described in~\cite{deb2013evolutionary}.
Algorithm~1 performs a non-dominated sort of individuals on different fronts and successively adds the individuals of $F_i$ to the set $S_t$ until their cardinality is greater than or equal to the desired population size $N$.
If $\vert S_t\vert = N$, the next generation $P_{t+1}$ inherits all individuals from $S_t$.
Otherwise, all previous fronts except the last front $F_i$ are added to the new generation.
To decide which $k$ individuals remaining in $F_i$ are included in $P_{t+1}$, the algorithm performs (1) normalization, (2) association, and (3) niching.
The three procedures are explained in detail in~\cite{deb2013evolutionary} and are briefly outlined below:
(1) First, the three values of the objective function of all individuals in $S_t$ are normalized.
(2) Then, the algorithm creates reference points $Z^f$ and associates each individual from $S_t$ with its nearest reference point.
(3) Finally, the niching procedure successively adds the individuals from $F_i$ to $P_{t+1}$ with which the fewest individuals are associated.
Niching ends when $k$ individuals have been added to the new generation $P_{t+1}$, so $\vert P_{t+1}\vert = N$ holds.
The memetic NSGA-III iterates until a termination criterion (e.g., generation or runtime limit) is satisfied.
The first front $F_1$ of the last generation is then the estimate of the Pareto front.

\section{Computational experiments}\label{sec:compex}

In this section, we present the computational experiments.
Section~\ref{ssec:compex-settings} describes the experimental setup of scheduling problems in the presence of real-world energy market data.
Sections~\ref{ssec:compex-makespan-cost} to \ref{ssec:compex-cost-co2} present the results and discuss the pairwise relationship of the objective function values, i.e., the relationship between makespan and energy cost, makespan and emissions, and energy cost and emissions.
Section~\ref{ssec:compex-implications} formulates implications.

\subsection{Instances and experimental setting} \label{ssec:compex-settings}

For the computational experiments, we use the benchmark set of~\cite{brandimarte1993routing}. Table~\ref{tab:benchmarks} shows the set that contains 15 instances for the FJSP with jobs and their respective operations as well as machines, making it suitable for emulating production schedules.
We select a 15-minute duration for each time step.
Thus, a schedule covers multiple periods with varying energy prices and emissions.

\begin{table}[h]
    \centering
    \caption{Benchmark instances by \citet{brandimarte1993routing}} \label{tab:benchmarks}
    \resizebox{0.6\textwidth}{!}{
	\begin{tabular}{lccccc}
        \toprule
        Instance&Jobs&Machines&\makecell{Operations\\per job}&\makecell{Operations\\in total}&\makecell{Time steps\\per operation}\\
        \midrule
        mk01&10&6&5-7&55&1-7\\ 
        mk02&10&6&5-7&58&1-7\\ 
        mk03&15&8&10&150&1-20\\ 
        mk04&15&8&3-10&90&1-10\\ 
        mk05&15&4&5-10&106&5-10\\ 
        mk06&10&10&15&150&1-10\\ 
        mk07&20&5&5&100&1-20\\ 
        mk08&20&10&5-10&225&5-20\\ 
        mk09&20&10&10-15&240&5-20\\ 
        mk10&20&15&10-15&240&5-20\\ 
        mk11&30&5&5-8&179&10-30\\ 
        mk12&30&10&5-10&193&10-30\\ 
        mk13&30&10&5-10&231&10-30\\ 
        mk14&30&15&8-12&277&10-30\\ 
        mk15&30&15&8-12&284&10-30\\ 
        \bottomrule
    \end{tabular}}
\end{table}

\begin{figure}[H]
    \centering
	\includegraphics[width=\textwidth]{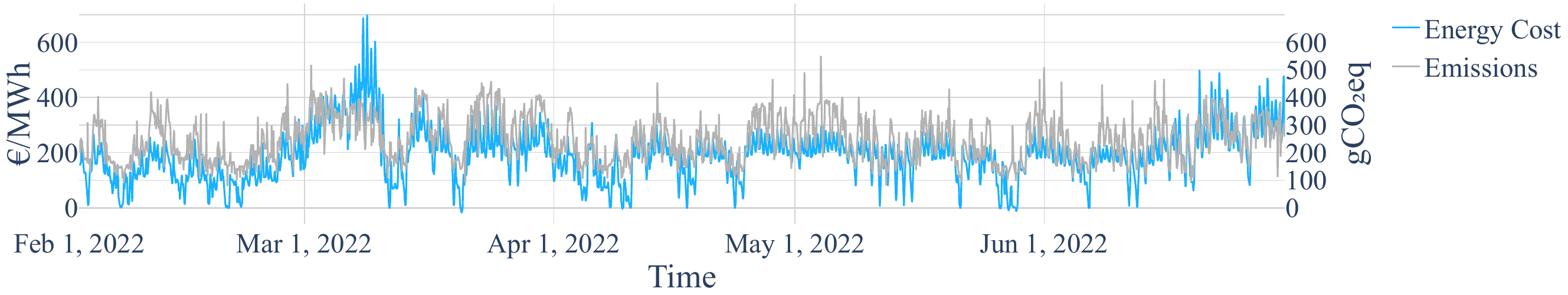}
	\caption{Electricity prices and emissions of the German energy market~\cite{smard2024}}
	\label{fig:energy-market}
\end{figure}

For consideration of energy consumption, we add an energy demand of 500$\frac{i}{\vert J\vert}$~kW for each operation of job $i\in J$, this means the demands of the operations range up to 500~kW.
We enrich the instances with real data from the German energy market published by the~\cite{smard2024}.
Figure~\ref{fig:energy-market} shows both electricity prices from the German wholesale market and emission values in hourly resolution from February 1\textsuperscript{st} to June 30\textsuperscript{th}, 2022.
Energy prices are determined by the European Energy Exchange.
Emission values are based on lifecycle emissions per generation technology, as outlined in~\cite{schlomer2014annex}.
The Pearson correlation coefficient for energy prices and energy emissions is 0.72.

For parameterization, we follow the settings of~\cite{burmeister2023memetic}.
We limit the runtime to 45 minutes to reflect the flexibility to respond to hourly price changes in the energy market.
The memetic NSGA-III is implemented in C\# 10 within the .NET 6 software framework.
The problem is solved on a Red Hat Enterprise Linux 8.5 (Oopta) operating system with an Intel Xeon Gold 6148 CPU, 20x2.4GHz, and 190 GByte main memory.

\subsection{Relation of makespan to energy cost}\label{ssec:compex-makespan-cost}

In this section, we focus on the relationship between makespan and energy cost without considering emissions.
The first three columns of Table~\ref{tab:res-msec} show the instance, the minimum makespan found by the memetic NSGA-III, and the associated energy cost.
The remaining columns show the percentage of energy cost that can be saved by increasing the makespan.

\begin{table}[b]
	\centering
	\caption{Energy cost savings (in \%) with increase in makespan}\label{tab:res-msec}    
    \resizebox{\textwidth}{!}{
	\begin{tabular}{lrr|rrrrp{0.25cm}|p{0.25cm}lrr|rrrr}
		\toprule
		& \multicolumn{2}{c}{\textbf{min}} &\multicolumn{4}{c}{\textbf{Increase of ms\textsuperscript{1} (\%)}} &&&& \multicolumn{2}{c}{\textbf{min}} &\multicolumn{4}{c}{\textbf{Increase of ms\textsuperscript{1} (\%)}}\\
  
		\textbf{Inst.} & \textbf{ms}\textsuperscript{1} & \textbf{ec}\textsuperscript{2} & \textbf{5} & \textbf{20} & \textbf{50} & \textbf{75}
        &&& \textbf{Inst.} & \textbf{ms}\textsuperscript{1} & \textbf{ec}\textsuperscript{2} & \textbf{5} & \textbf{20} & \textbf{50} & \textbf{75}\\
		\midrule
		mk01 & 42 & 3965 & 2.0 & 6.5 & 17.0 & 22.3 && & mk09 & 341 & 45041 & 5.5 & 16.7 & 32.5 & 38.6\\
		mk02 & 29 & 3204 & 0.1 & 2.2 & 2.3 & 3.1 & && mk10 & 263 & 40158 & 7.0 & 10.2 & 16.1 & 26.3\\
		mk03 & 204 & 13954 & 1.7 & 2.5 & 2.5 & 2.5 && & mk11 & 621 & 41038 & 8.6 & 10.1 & 10.1 & 10.1\\
		mk04 & 66 & 6943 & 0.7 & 7.7 & 26.5 & 51.7 && & mk12 & 524 & 49137 & 7.7 & 18.6 & 18.6 & 18.6\\
		mk05 & 174 & 11791 & 5.0 & 6.9 & 6.9 & 6.9 & && mk13 & 457 & 71059 & 11.8 & 30.4 & 36.7 & 36.7\\
		mk06 & 77 & 7988 & 3.7 & 15.9 & 45.0 & 49.3 && & mk14 & 694 & 63382 & 2.2 & 2.2 & 2.2 & 2.2\\
		mk07 & 144 & 10658 & 14.6 & 15.9 & 15.9 & 15.9 && & mk15 & 429 & 89821 & 8.4 & 24.3 & 36.6 & 36.6\\
		mk08 & 523 & 37481 & 8.9 & 14.5 & 14.5 & 14.5 & &&  \multicolumn{4}{c}{}  \\		
		\bottomrule
		\multicolumn{7}{l}{\rule{0pt}{2ex}\textsuperscript{1} Makespan (time steps), \textsuperscript{2} Energy cost (\euro)}\\	
	\end{tabular}}
\end{table}

A 5\% makespan increase saves between 0.1\% (mk02) to 14.6\% (mk07) of energy cost across all instances.
For a 20\% makespan increase, the savings range from 2.2\% (mk02 and mk14) to 30.4\% (mk13).
For increases in makespan of 50\% and 75\%, the energy cost remain stagnant for instances mk03, mk05, mk08, mk11, mk12, and mk14, showing no further improvements.
For other instances, energy cost continues to decrease, achieving savings of up to 45.0\% (mk06).
At a 75\% makespan increase, instances mk06 and mk04 show the highest energy cost savings with 49.3\% and 51.7\%, while instances mk03 and mk14 show the lowest savings (2.5\% and 2.2\%).

The average savings in energy cost are 5.86\%, 12.31\%, 18.89\%, and 22.35\% for makespan increases of 5\%, 20\%, 50\%, and 75\%, respectively.
The results indicate that small increases in makespan are more efficient for energy cost savings, as the relative increase in savings surpasses the increase in makespan for 9 out of 15 instances.
However, efficiency decreases with larger increases in makespan.
With a 20\% increase in makespan, the savings in energy cost exceed the 20\% mark in 2 out of 15 instances.
Savings with an increase in makespan of 50\% and 75\% are below 50\% and 75\%, respectively, for all instances.

\subsection{Relation of makespan to emissions}\label{ssec:compex-makespan-co2}

Table~\ref{tab:res-msem} shows the results of the reduction in emissions with increasing makespan and follows the structure of Table~\ref{tab:res-msec}.
A 5\% makespan increase saves between 0.4\% (mk02) and 10.5\% (mk13) of the emissions.
With a 20\% makespan increase, the savings range from 1.0\% (mk14) to 19.1\% (mk13).
For makespan increases of 50\% and 75\%, the emissions remain constant for the instances mk03, mk05, mk07, mk08, mk11, and mk14.
Other instances show emissions savings of up to 23.7\% (mk15) and 24.1\% (mk09) with makespan increases of 50\% and 75\%, respectively.

\begin{table}[ht]
	\centering
	\caption{Emissions savings (in \%) with increase in makespan}\label{tab:res-msem}
    \resizebox{\textwidth}{!}{
	\begin{tabular}{lrr|rrrrp{0.25cm}|p{0.25cm}lrr|rrrr}
		\toprule
		& \multicolumn{2}{c}{\textbf{min}} &\multicolumn{4}{c}{\textbf{Increase of ms\textsuperscript{1} (\%)}} &&&& \multicolumn{2}{c}{\textbf{min}} & \multicolumn{4}{c}{\textbf{Increase of ms\textsuperscript{1} (\%)}}\\
		\textbf{Inst.} & \textbf{ms}\textsuperscript{1} & \textbf{em}\textsuperscript{2} & \textbf{5} & \textbf{20} & \textbf{50} & \textbf{75} 
        &&& \textbf{Inst.} & \textbf{ms}\textsuperscript{1} & \textbf{em}\textsuperscript{2} & \textbf{5} & \textbf{20} & \textbf{50} & \textbf{75}\\
		\midrule
		mk01 & 42 & 5.215 & 1.3 & 5.4 & 12.7 & 17.7 &&& mk09 & 341 & 68.240 & 3.4 & 11.8 & 20.4 & 24.1 \\
        mk02 & 29 & 4.627 & 0.4 & 2.2 & 3.7 & 5.3 &&& mk10 & 263 & 61.328 & 5.0 & 7.6 & 13.4 & 21.2 \\
        mk03 & 204 & 24.257 & 2.9 & 2.9 & 2.9 & 2.9 &&& mk11 & 621 & 73.575 & 4.7 & 5.6 & 5.6 & 5.6 \\
        mk04 & 66 & 9.370 & 1.0 & 7.9 & 14.9 & 17.0 &&& mk12 & 524 & 83.272 & 5.7 & 10.4 & 10.7 & 10.7 \\
        mk05 & 174 & 18.303 & 3.5 & 3.6 & 3.6 & 3.6 &&& mk13 & 457 & 109.633 & 10.5 & 19.1 & 23.1 & 23.1 \\
        mk06 & 77 & 10.436 & 3.2 & 11.2 & 14.7 & 17.6 &&& mk14 & 694 & 118.605 & 1.0 & 1.0 & 1.0 & 1.0 \\
        mk07 & 144 & 17.517 & 6.3 & 7.1 & 7.1 & 7.1 &&& mk15 & 429 & 139.310 & 6.9 & 17.6 & 23.7 & 23.7 \\
        mk08 & 523 & 64.146 & 5.4 & 8.3 & 8.3 & 8.3 &&& \multicolumn{4}{c}{}\\
		\bottomrule
		\multicolumn{7}{l}{\rule{0pt}{2ex}\textsuperscript{1} Makespan (time steps), \textsuperscript{2} Emissions (tons of CO\textsubscript{2}eq)}\\	
	\end{tabular}}
\end{table}

The average emissions savings are 4.08\%, 8.11\%, 11.05\%, and 12.59\% for relative makespan increases of 5\%, 20\%, 50\%, and 75\%, respectively.
Again, the most efficient savings are observed for smaller instances.
For a 5\% makespan increase, the relative increase in savings exceeds the increase in makespan for 6 of 15 instances.

The emissions savings are lower than the energy cost savings in Table~\ref{tab:res-msec}.
This can be attributed to the fact that the market values of the emissions have a positive value range, while the energy cost may have negative values.
As a result, the algorithm has fewer opportunities to schedule jobs with high energy demands at favorable times when considering emissions.

\subsection{Relation of energy cost to emissions}\label{ssec:compex-cost-co2}
Table~\ref{tab:res-ecem} shows the results of the emissions savings as energy cost increase without considering the makespan.
It is structured similarly to Table~\ref{tab:res-msec} and \ref{tab:res-msem}.
For instance mk02, the absolute value of the energy cost was used because the instance has few jobs that are processed for negative energy cost.

\begin{table}[b]
	\centering
	\caption{Emissions savings (in \%) with increase in energy cost}\label{tab:res-ecem}
	\resizebox{\textwidth}{!}{
    \begin{tabular}{lrr|rrrrp{0.25cm}|p{0.25cm}lrr|rrrr}
		\toprule
		& \multicolumn{2}{c}{\textbf{min}} &\multicolumn{4}{c}{\textbf{Increase of ec\textsuperscript{1} (\%)}} &&&& \multicolumn{2}{c}{\textbf{min}} &\multicolumn{4}{c}{\textbf{Increase of ec\textsuperscript{1} (\%)}}\\
        \textbf{Inst.} & \textbf{ec}\textsuperscript{1} & \textbf{em}\textsuperscript{2} & \textbf{5} & \textbf{20} & \textbf{50} & \textbf{75} 
        &&&\textbf{Inst.} & \textbf{ec}\textsuperscript{1} & \textbf{em}\textsuperscript{2} & \textbf{5} & \textbf{20} & \textbf{50} & \textbf{75}\\
		\midrule
        mk01 & 5 & 3.249 & 0.3 & 1.7 & 5.4 & 6.2 &&&mk09 & 10170 & 45.452 & 2.0 & 5.7 & 6.1 & 6.1 \\
        mk02\textsuperscript{3} & -2 & 3.098 & 2.1 & 6.6 & 9.4 & 14.2 &&&mk10 & 7182 & 40.082 & 1.8 & 5.0 & 6.7 & 7.3 \\
        mk03 & 1282 & 17.464 & 1.3 & 6.6 & 8.1 & 9.7 &&&mk11 & 18933 & 60.929 & 1.4 & 2.5 & 2.5 & 2.5 \\
        mk04 & 71 & 6.343 & 1.6 & 5.7 & 8.0 & 8.9 &&&mk12 & 16571 & 62.976 & 1.0 & 2.7 & 2.8 & 2.8 \\
        mk05 & 838 & 13.940 & 2.2 & 4.0 & 5.2 & 6.6 &&&mk13 & 22309 & 74.076 & 3.4 & 3.6 & 3.6 & 3.6 \\
        mk06 & 132 & 7.506 & 0.2 & 8.6 & 12.1 & 13.5 &&&mk14 & 32770 & 100.836 & 1.4 & 2.2 & 2.2 & 2.2 \\
        mk07 & 668 & 13.517 & 5.6 & 6.9 & 10.7 & 11.5 &&&mk15 & 29438 & 93.441 & 3.3 & 4.6 & 4.6 & 4.6 \\
        mk08 & 13019 & 48.015 & 0.7 & 1.4 & 1.4 & 1.4 &&&\multicolumn{4}{c}{}\\
		\bottomrule		
		\multicolumn{16}{l}{\rule{0pt}{2ex}\textsuperscript{1} Energy cost (\euro), \textsuperscript{2} Emissions (tons of CO\textsubscript{2}eq), \textsuperscript{3} Since the value is negative, we use}\\
        \multicolumn{16}{l}{\rule{0pt}{2ex}\quad the absolute value for the percentage increase.}\\
	\end{tabular}}
\end{table}

A 5\% energy cost increase saves between 0.7\% (mk08) and 5.6\% (mk06) of emissions.
For increases in energy costs of 20, 50 and 75\%, the maximum savings are 8.6 (mk06), 12.1 (mk06), and 14.2\% (mk02), respectively, and the minimum savings are 1.4\% (mk08).
Except for the 5.6\% reduction in emissions for a 5\% deterioration in energy cost for the mk07 instance, the relative savings are less than the relative deterioration in all other cases.
The average emissions savings are 1.89\%, 4.52\%, 5.92\%, and 6.74\% for relative energy cost increases of 5\%, 20\%, 50\%, and 75\%, respectively.

\subsection{Implications} \label{ssec:compex-implications}
\begin{figure}[h]
    \centering
	\includegraphics[width=.7\linewidth]{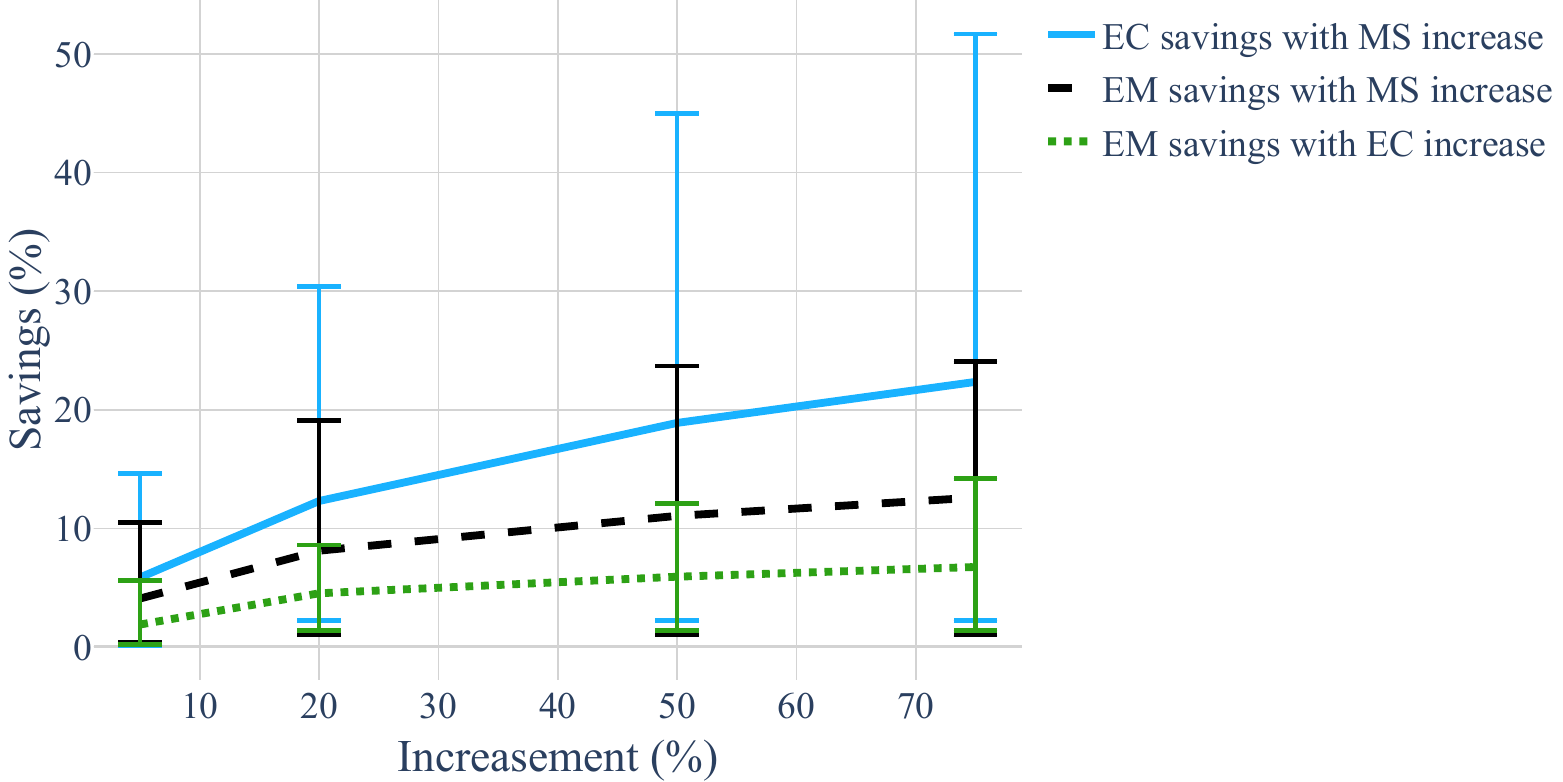}
	\caption{Minimum, mean and maximum savings}
	\label{fig:tradeoff}
\end{figure}

In this section, we draw implications from the results.
Figure~\ref{fig:tradeoff} summarizes the medians of the percentage savings in energy costs when increasing the makespan (solid line), emissions when increasing the makespan (dashed line), and emissions when increasing the energy cost (dotted line).
The error bars show the minimum and maximum values.

First, extending the makespan can lead to savings in energy costs and emissions.
The percentage values observed in Figure~\ref{fig:tradeoff} indicate disproportionately high average savings in energy costs when the makespan increases by 5\%.
In contrast, increases in makespan beyond 5\% and reductions in emissions show lower percentage savings compared to the initial increase in makespan.
Energy costs show a higher savings potential than emissions, as the energy market can have negative prices, while emissions cannot be negative. 
Although an increase in makespan results in a disproportionately low increase in the percentage of savings in energy costs and emissions, the absolute savings remain considerable.
This is because the underlying absolute energy costs range between \euro3,204 and \euro89,821, and emission values range from 4.63 to 139.31 tons of CO\textsubscript{2}eq.
We advise decision-makers to consider the benefits of enhancing production flexibility for potential savings.

Second, opting for an increase in energy cost to reduce emissions is less effective than increasing makespan.
The average emission reductions achieved by creating time flexibility with a longer makespan is higher than that resulting from the acceptance of higher energy cost for the purchase of renewable energy.
If decision-makers are unable to consider an increase in makespan (e.g., due to deadlines), it may still be attractive for manufacturers to balance higher energy costs against emissions savings.
Additional costs for reducing or avoiding emissions could be incorporated into adjusted product pricing, thereby promoting sustainable production without extending the makespan.

\section{Conclusion} \label{sec:conclusion}
Our study addresses the challenge of managing flexible loads in manufacturing industries, with the aim of optimizing schedules with respect to makespan, energy cost, and emissions.
Drawing from existing literature, we introduce a linear optimization model for a Green FJSP and develop a memetic NSGA-III approach as a three-objective solution method.
Evaluating our method on benchmark instances enriched with real energy market data from Germany, we analyze potential savings in energy costs and emissions.
This helps production planners assess the benefits of flexible production planning adapted to the energy market.
We find that even a small extension of the makespan can lead to substantial savings in energy costs and emissions, with potential savings in energy costs being higher than those in emissions.
In addition, emissions can be reduced at the expense of higher energy costs.
However, while the potential savings from increased makespan are significant, they do not match the extent of savings achievable through flexible production.
Our findings underscore the importance of considering the interplay between the considered objectives to strengthen sustainable production planning in manufacturing companies.

Based on the limitations and findings of this study, we advise avenues for further research.
First, we recommend comparing the solutions obtained from our approach with those of other state-of-the-art methods.
Given that this work focuses on exploring trade-offs between objectives, our results currently serve as lower bounds for potential savings, suggesting that improved solutions may also be possible.
In future statistical analysis, solution quality could be examined by assessing differences in efficiency and solution quality, as well as calculating an optimality gap. 

Second, we recommend extending our approach to account for data uncertainty, as the current model assumes deterministic knowledge of energy costs and emissions.
In a dynamic variant, schedules could be developed using energy market data and forecasts, allowing real-time adjustments to accommodate evolving grid conditions and renewable energy availability.
Investigating a dynamic model would provide insights into how decision-makers can adapt to fluctuations in the energy market.

Third, we recommend that our approach be analyzed on real production data.
While we use real energy market data combined with scheduling problems from a benchmark set to analyze savings potentials for different-sized problem cases, we also recommend evaluating our approach with production orders from a real manufacturer.
In this way, our goal is to achieve a better understanding of the trade-off between makespan, energy cost, and emissions.

\section*{Acknowledgments}
The authors gratefully acknowledge the financial support provided by the state of North Rhine-Westphalia, Germany, as part of the \textit{progres.nrw} program area, in the framework of Re$^2$Pli (project number EFO 0127A) and the funding of this project by computing time provided by the Paderborn Center for Parallel Computing (PC$^2$).

\bibliographystyle{agsm}  
\bibliography{references}

\end{document}